\documentclass[11pt,a4paper]{article}
\usepackage{times,latexsym}
\usepackage{url}
\usepackage[T1]{fontenc}
\usepackage{booktabs}
\usepackage{amsmath}
\usepackage{amssymb}
\usepackage{graphicx}
\usepackage{multirow}
\usepackage{arydshln}           
\usepackage{caption}            

\usepackage[acceptedWithA]{tacl2021v1}

\iftaclpubformat 

\else

\fi

\title{Corrections Meet Explanations: A Unified Framework for Explainable Grammatical Error Correction}

\author{
    \textbf{Jingheng Ye}\textsuperscript{1}\thanks{~~Equal contribution.}~~,
    \textbf{Shang Qin}\textsuperscript{1}\footnotemark[1]~~,
    \textbf{Yinghui Li}\textsuperscript{1},\\
    \textbf{Hai-Tao Zheng}\textsuperscript{1,2}\thanks{~Corresponding author: Hai-Tao Zheng. \texttt{(Email: zheng.haitao@sz.tsinghua.edu.cn)}}~~, 
    \textbf{Shen Wang}\textsuperscript{4},
    \textbf{Qingsong Wen}\textsuperscript{4} \\
    \textsuperscript{1}Tsinghua University,
    \textsuperscript{2}Peng Cheng Laboratory, \\
    \textsuperscript{3}Squirrel Ai Learning \\
    \texttt{\{yejh22,qin-s23\}@mails.tsinghua.edu.cn}
}

\date{}
\newcommand{\Method}{EXGEC}
\newcommand{\Dataset}{EXPECT-\textit{denoised}}

\begin{document}
\maketitle
\begin{abstract}

Grammatical Error Correction (GEC) faces a critical challenge concerning explainability, notably when GEC systems are designed for language learners. Existing research predominantly focuses on explaining grammatical errors extracted in advance, thus neglecting the relationship between explanations and corrections. To address this gap, we introduce \textbf{EXGEC}, a unified explainable GEC framework that integrates explanation and correction tasks in a generative manner, advocating that these tasks mutually reinforce each other. Experiments have been conducted on EXPECT, a recent human-labeled dataset for explainable GEC, comprising around 20k samples. Moreover, we detect significant noise within EXPECT, potentially compromising model training and evaluation. Therefore, we introduce an alternative dataset named \Dataset{}, ensuring a more objective framework for training and evaluation. Results on various NLP models (BART, T5, and Llama3) show that EXGEC models surpass single-task baselines in both tasks, demonstrating the effectiveness of our approach.

\end{abstract}

\section{Introduction}\label{sec:intro}
Writing proficiently poses significant challenges for language learners who often struggle to produce grammatically correct and coherent texts~\citep{li2024rethinking,DBLP:conf/acl/LiZLLLSWLCZ22,DBLP:conf/emnlp/YeLZLM0023}. Therefore, GEC systems~\citep{bryant2023grammatical} are developed to detect and rectify all grammatical errors in texts~\citep{ye-etal-2023-mixedit,huang-etal-2023-frustratingly}. Research advancements in GEC encompass multi-language~\citep{rothe-etal-2021-simple,DBLP:journals/corr/abs-2307-09007,DBLP:conf/emnlp/MaLSZHZLLLCZS22}, multi-modality~\citep{fang-etal-2023-improving,DBLP:conf/emnlp/LiMZLLHLLC022,DBLP:conf/acl/LiXC0LMJLZZS24}, and document-level~\citep{yuan-bryant-2021-document,DBLP:conf/emnlp/DuW0D0LZVZSZGL024}.

However, the explainability of GEC~\citep{hanawa-etal-2021-exploring,DBLP:journals/corr/abs-2407-00924,DBLP:journals/corr/abs-2407-00934} remains under-explored due to its intrinsic difficulties. Given that neural GEC systems generally function as intricate black-box models, their internal processes are not transparent~\citep{zhao2023explainability}. The absence of explainability can result in inadequacies in educational scenarios, where L2 learners might find it difficult to completely understand outputs from GEC systems without knowing the rationale behind corrections. Providing corrections with explanations fosters appropriate trust by clarifying the linguistic principles and logical mechanisms underpinning model decisions in a comprehensible way, thereby aiding educationally K-12 students and L2 speakers~\citep{bitchener2005effect,sheen2007effect,DBLP:journals/eswa/LiMCHHLZS25,ye2025position}. Moreover, explainability offers insights that help identify unintentional biases and risks for researchers, functioning as a debugging tool to enhance model performance~\citep{ludan-etal-2023-explanation,DBLP:conf/icassp/ZhangLZMLCZ23}.

\begin{figure*}[t]
\centering
\includegraphics[scale=0.31]{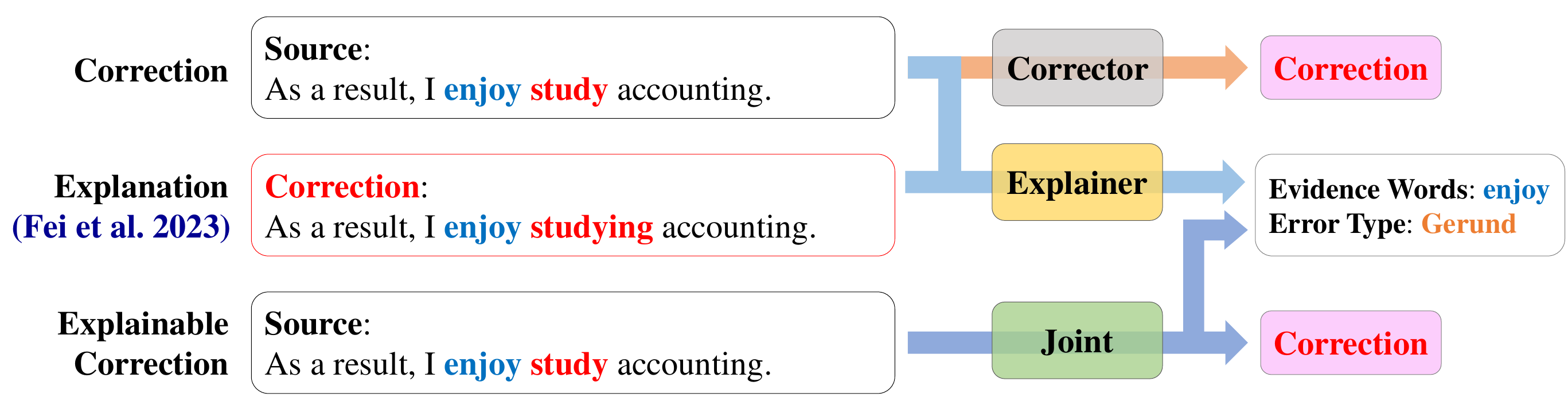}
\caption{Comparison between tasks of correction, explanation~\citep{fei-etal-2023-enhancing}, and explainable GEC.}
\label{fig:intro}
\end{figure*}

The paper focuses on EXPECT~\citep{fei-etal-2023-enhancing}, an explainable GEC dataset characterized by human-labeled \textit{evidence words} and \textit{grammatical error types} annotations, designed to assist language learners in understanding the corrections from GEC systems. These evidence words, referred to as extractive rationales\footnote{Following EXPECT~\citep{fei-etal-2023-enhancing}, we use the term ``evidence words'' throughout the paper except in Section~\ref{sec:related_works}.}, provide precise cues for corrections, thereby enabling learners to comprehend the rationale underlying the corrections. The error types within EXPECT encompass 15 categories grounded in pragmatism~\citep{skehan1998cognitive,shichungui}, facilitating learners in inferring abstract grammatical rules from particular errors through inductive reasoning. However, existing studies~\citep{song-etal-2024-gee} primarily concentrate on post hoc explanation, neglecting the interaction between the explanation and correction tasks as represented in Figure~\ref{fig:intro}.

To explore the interaction between explanation and correction tasks, we introduce \textbf{EXGEC} (\textbf{EX}plainable \textbf{G}rammatical \textbf{E}rror \textbf{C}orrection), a unified multi-task explainable GEC framework that formulates the multi-task problem as a generative task. The framework can jointly correct ungrammatical sentences, extract evidence words, and classify grammatical errors~\cite{zou2025revisiting} in different prediction orders within an architecture. Our research indicates that learning correction and explanation tasks together can be mutually beneficial and the prediction orders affect the task performance. More specifically, pre-explaining models achieve better correction performance but lower explanation performance compared to post-explaining models. Nevertheless, both models show improved or comparable correction and explanation performance compared to their respective baselines.

Moreover, we find that EXPECT is not an ideal dataset for explainable GEC. This is due to the presence of numerous unidentified grammatical errors in EXPECT, which would disturb the extraction of evidence words and the prediction of grammatical errors. Therefore, it will lead to a bias in the training and evaluation process. Consequently, we reconstruct EXPECT to correct the unidentified errors while ensuring each sentence contains only one distinct error~\citep{fei-etal-2023-enhancing}. The resulting dataset is called \Dataset{}. By training and evaluating EXGEC models on our proposed \Dataset{}, we can obtain unbiased results reflecting their real abilities in both the correction and the explanation tasks. In summary, our contributions are three folds:

\begin{itemize}
\item [(1)] We present EXGEC, a comprehensive framework that integrates correction and explanation components. This adaptable design facilitates the investigation of the interplay between correction and explanation tasks when utilizing various prediction sequences.

\item [(2)] We recognize a potential critical limitation in EXPECT and reconstruct it into \Dataset{}, thereby enhancing the training and evaluation framework for EXGEC models.

\item [(3)] We perform extensive experiments employing three language models (BART, T5, and Llama3) to demonstrate the beneficial interaction between the two tasks and substantiate the efficacy of our approach.

\end{itemize}


\section{Related Work}
\label{sec:related_works}

\subsection{Explainable GEC}
Many GEC systems focus solely on correction without offering explanations~\citep{davis-etal-2024-prompting,ye2022focus,ye2023system}. To address this limitation, recent research has investigated multiple techniques to enhance the explainability of GEC. One technique is Example-based GEC~\citep{kaneko-etal-2022-interpretability,vasselli-watanabe-2023-closer}, which boosts explainability by retrieving examples that are similar to the input instance based on defined grammar rules. GEE~\citep{song-etal-2024-gee} develops a two-step pipeline for GEC explanation generation.~\citet{kaneko-okazaki-2024-controlled} explore the generation of natural language explanations by prompting large language models (LLMs). Another relevant task is feedback comment generation (FCG)~\citep{nagata-2019-toward,nagata-etal-2021-shared, hanawa-etal-2021-exploring}, which aims to automatically create feedback comments, like hints or explanatory notes, to aid in writing learning. However, it suffers from expensive costs associated with data annotation~\citep{nagata2020creating}. Furthermore, it is often explored with limited access to only a subset of English grammatical error types due to the complexity of the task~\citep{nagata-2019-toward}.

Despite these efforts, no research has comprehensively examined the positive interaction between correction and explanation tasks during training. In contrast, our work focuses on studying whether learning a multi-task model can outperform the respective single-task models.

\subsection{Learning with Explanations}
Explainability of NLP tasks is a critical research direction and has been given serious attention recently, especially due to the ``black box'' nature of LLMs~\citep{dalal-etal-2024-inference,hu2024learning,yu2024mind,saeed2024sumex,DBLP:conf/aaai/YuJLHWLCLLTZZXH24,DBLP:conf/coling/XuLD0CJZLXH25}. Prior studies have shown that training models to produce task predictions and explanations concurrently can boost performance in vision-language tasks~\citep{majumder2022knowledge,DBLP:conf/icassp/LiCLXCZ23,DBLP:conf/aaai/LiL0LHYY024} and several downstream NLP tasks, including text classification~\citep{li2022unifying,DBLP:conf/nips/LiZLML0HY24}, commonsense reasoning~\citep{veerubhotla-etal-2023-shot,yan2025position,DBLP:conf/coling/HuangMLHZ0Z24,DBLP:conf/sigir/LiLHYS022,DBLP:journals/tkde/LiHZZLLCZS23}, and complaint detection~\citep{singh-etal-2023-peeking}. An essential aspect of this research is the development of self-Rationalization models that generate task predictions along with corresponding explanations to enhance the explainability or task performance of neural networks. There are two main methods for building self-Rationalization models: 1) \textit{extracting key input tokens responsible for task predictions}, referred to as extractive rationales~\citep{deyoung-etal-2020-eraser}, and 2) \textit{creating natural language explanations}~\citep{narang2020wt5}, which serve as a natural interface between models and human users. To refine the performance and trustworthiness of Seq2Seq models,~\citet{lakhotia-etal-2021-fid} developed an extractive fusion-in-decoder architecture within the ERASER benchmark~\citep{deyoung-etal-2020-eraser}, a well-known benchmark for rationale extraction across various datasets and tasks.~\citet{li2022unifying} introduced a combined text classification and rationale extraction model to improve explainability and robustness. Recognizing the synergy between extractive rationales and natural language explanations,~\citet{majumder2022knowledge} integrated both components into a self-Rationalization framework.

\paragraph{Explanation-augmented knowledge distillation.}
Leveraging in-context learning~\citep{brown2020language} and the chain-of-thought (CoT) reasoning~\citep{chu2023survey} of LLMs, many recent studies employ the natural language explanations produced by LLMs with chain-of-thought prompting~\citep{lampinen-etal-2022-language,li2023symbolic} to enhance the development of smaller reasoning models using knowledge distillation~\citep{zhang2024elad}, thereby boosting task performance~\citep{li2024explanations,ho-etal-2023-large,hsieh-etal-2023-distilling} or improving faithfulness~\citep{wang-etal-2023-scott}.
However, convincing and wrong explanations generated by LLMs can foster unwarranted confidence in tackling NLP tasks~\citep{madsen-etal-2024-self,pruthi-etal-2022-evaluating}, particularly in educational contexts emphasizing faithfulness~\citep{lyu-etal-2024-towards} and correctness~\citep{huang-etal-2024-chatgpt}. Consequently, this paper emphasizes model training with human-annotated datasets.

\section{Motivation and Methodology}
\label{sec:method}

\subsection{Problem Definition}
\label{subsec:problem_definition}
The objective of this study is to simultaneously address the correction and explanation tasks using a Seq2Seq-based generation approach. Specifically, given an ungrammatical input sentence $X = \{x_0, x_1, \cdots, x_n\}$, where $n$ represents the length of the input sentence, the joint models are designed to learn both tasks. The correction task involves converting the ungrammatical input into a grammatical output $Y = \{y_0, y_1, \cdots, y_m\}$, where $m$ denotes the length of the output. The explanation task is divided into two sub-tasks: 1) \textbf{classifying} grammatical errors, and 2) \textbf{extracting} evidence words. For the classification sub-task, the joint models are required to output a grammatical error type label $c$ ($c \in C$), where $C$ consists of 15 possible grammatical error type classes as defined in EXPECT. For the extraction sub-task, the models must extract evidence words $E(X) = \{e_0, e_1, \cdots, e_k\} \subset X$ that provide clear and comprehensive clues for corrections.

\begin{figure*}[!t]
\centering
\includegraphics[scale=0.31]{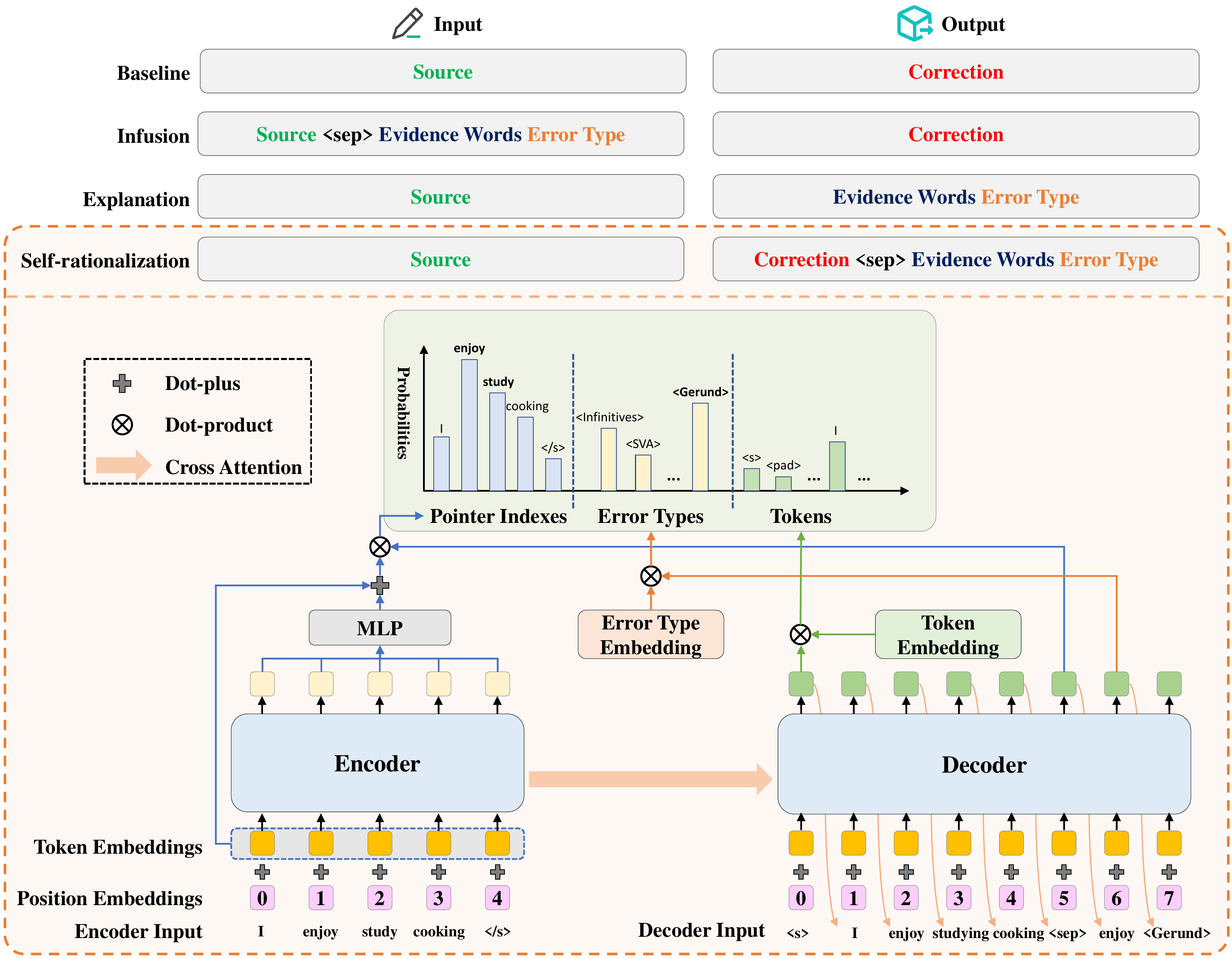}
\caption{(a) Top block: Four settings compatible with EXGEC.
(b) Overview of our \textit{Self-Rationalization} model. The decoder can 1) output \textit{corrections} from the model's token vocabulary, 2) extract \textit{evidence words} as source indexes by leveraging the pointer mechanism, and 3) predict an \textit{error type} from the predefined set of error type classes. ``<s>'' and ``</s>'' are start-of-sentence and end-of-sentence tokens. Although we show the case in a Seq2Seq model, other generative models like Llama3 can work similarly.}
\label{fig:overview}
\end{figure*}

\subsection{Explainable GEC as Generation Task}
\label{subsec:model}
We introduce four distinct training settings to examine the interaction between explanation and correction tasks during the training phase, as illustrated in Figure~\ref{fig:overview}. These configurations are as follows: 1) the absence of explanations (\textit{Baseline}), which represents the conventional GEC setup, 2) the integration of explanations as supplementary input (\textit{Infusion}), 3) the generation of explanations as output (\textit{Explanation}), and 4) the inclusion of explanations as an auxiliary output (\textit{Self-Rationalization}).

This paper presents EXGEC, a unified generative framework designed for explainable grammatical error correction, which seamlessly integrates multiple settings within a singular architecture. In the Infusion setting, a special token, ``<sep>'', acts as a delimiter between the source sentence and the ensuing explanation, comprising evidence words and the identified error type. Conversely, in the Explanation setting, the model derives an explanation based solely on the information provided by the source sentence. Within the Self-Rationalization paradigm, the models concurrently generate both a correction and an explanation. The Self-Rationalization setting is depicted in detail in Figure~\ref{fig:overview}. Other settings can be adapted with minimal modifications. Moreover, the sequence in which corrections and explanations are predicted can be altered, facilitating a deeper understanding of task interactions.

We first describe how our EXGEC addresses tasks in a unified generative framework within the Self-Rationalization setting. Utilizing a pointing mechanism~\citep{vinyals2015pointer}, EXGEC can identify evidence words by directly generating the source indices of an ungrammatical sentence, thus eliminating the possibility of generating invalid evidence words. Given an ungrammatical source sentence $X$, the encoder translates $X$ into a hidden representation $\mathbf H$ as follows:
\begin{equation}
\mathbf{H}^e = \operatorname{Encoder}(X),
\end{equation}
\noindent where $\mathbf{H}^e\in\mathbb{R}^{n\times d}$, and $d$ is the hidden size. At each time step $t$, the decoder produces the hidden state $\mathbf{h}^d_t\in\mathbb{R}^d$ based on the previous output sequence $\hat{Y}_{<t}$, which is computed as follows:
\begin{equation}
\mathbf{h}^d_t = \operatorname{Decoder}(\mathbf{H}^e,\hat{Y}_{<t}).
\end{equation}

Next, the hidden state $\mathbf{h}^d_t\in\mathbb{R}^d$ is utilized to compute three forms of logits: 1) \textit{correction token logits}, responsible for the correction phase~\citep{vaswani2017attention}, 2) \textit{evidence pointer logits}, which determine the probabilities of source indices for evidence extraction, and 3) \textit{error type logits}, used for classifying error types.
Drawing inspiration from~\citet{yan-etal-2021-unified-generative}, we add an extra MLP layer~\cite{DBLP:journals/patterns/LiuLTLZ22} and calculate the probability distribution $P_t$ as follows:
\allowdisplaybreaks
\begin{align}
\mathbf{E}^e &= \operatorname{TokenEmbed}(X) \in \mathbb{R}^{n\times d}, \\[10pt]
\bar{\mathbf{H}}^e &= \alpha \mathbf{E}^e + (1-\alpha) \operatorname{MLP}(\mathbf{H}^e) \in \mathbb{R}^{n\times d}, \\[10pt]
\mathbf{V}^d &= \operatorname{TokenEmbed}(V) \in \mathbb{R}^{|V|\times d}, \\[10pt]
\mathbf{C}^d &= \operatorname{TypeEmbed}(C) \in \mathbb{R}^{|C|\times d}, \\[10pt]
P_t &= \operatorname{softmax}([\mathbf{V}^d \otimes \mathbf{h}^d_t; \bar{\mathbf{H}}^e \otimes \mathbf{h}^d_t; \mathbf{C}^d \otimes \mathbf{h}^d_t]),
\end{align}
\noindent where TokenEmbed refers to the embeddings that are shared between the encoder and decoder, $\alpha \in \mathbb{R}$ is a hyper-parameter responsible for balancing the trade-off between embeddings and encoder hidden representation, $V$ represents the token vocabulary, $[\cdot \ ; \ \cdot]$ denotes the concatenation operation in the first dimension, the symbol $\otimes$ means the dot product operation, and
$P_t\in\mathbb{R}^{|V| + n + |C|}$ represents the probability distribution at the current time step $t$.

The pointer index cannot be directly inputted into the decoder. So we utilize the Index2Token conversion to transform the indexes into tokens~\citep{yan-etal-2021-unified-generative}. Furthermore, the sequence of generating corrections and explanations can be restructured, potentially offering valuable insights into the deeper understanding of the interaction between both tasks. The probability distribution of Baseline and Infusion models is restricted to the token vocabulary. Conversely, the probability distribution of Explanation models is confined to the combination of pointer indexes and error types.

\subsection{Loss Weighting}
\label{subsec:loss_weighting}

Taking into account the heterogeneity of correction and explanation tasks, we construct the overall loss function in the form of a weighted sum, which is defined as follows:
\begin{equation}\begin{small}\begin{aligned}
\mathcal{L} &= \mathcal{L}_{cor} + \lambda \cdot \mathcal{L}_{exp} \\
&= -\sum^m_{i=0} \Big[ \mathbb{I}(y_i \in V)\log p_i + \lambda \mathbb{I}(y_i \not\in V) \log p_i \Big],
\end{aligned}\label{eq:loss}\end{small}\end{equation}
\noindent where $\lambda$ is responsible for balancing both tasks and $\mathbb{I}$ is the indicator function. During the inference stage, we generate the entire target sequence in an autoregressive manner and then separate different parts from the target.

\begin{table*}[tp!]
\centering
\scalebox{0.80}{
\begin{tabular}{ll}
\toprule

\textbf{W\&I+LOCNESS Source}  &
However \textcolor{blue}{I} sometimes do a skipping to \textcolor{blue}{fit myself} .   \\

\textbf{W\&I+LOCNESS Target}  &
However \textcolor{red}{,} I sometimes do skipping to \textcolor{red}{keep myself fit} .  \\

\hdashline

\textbf{EXPECT Source}  &
However \textcolor{blue}{I} sometimes do skipping to \textcolor{blue}{fit myself} .    \\

\textbf{EXPECT Target}  &
However \textcolor{blue}{I} sometimes do skipping to \textcolor{red}{keep myself fit} .    \\

\hdashline

\textbf{\Dataset{} Source}  &
However \textcolor{red}{,} I sometimes do skipping to \textcolor{blue}{fit myself} .   \\

\textbf{\Dataset{} Target}  &
However \textcolor{red}{,} I sometimes do skipping to \textcolor{red}{keep myself fit} .    \\

\midrule

\textbf{W\&I+LOCNESS Source}  &
\textcolor{blue}{i} have a dog \textcolor{blue}{it} name 's \textcolor{blue}{chente , it} is a golden \textcolor{blue}{retriver} .   \\

\textbf{W\&I+LOCNESS Target}  &
\textcolor{red}{I} have a dog \textcolor{red}{and its} name 's \textcolor{red}{Chente . It} is a golden \textcolor{red}{retriever} .  \\

\hdashline

\textbf{EXPECT Source}  &
\textcolor{blue}{i} have a dog \textcolor{blue}{its} name 's \textcolor{blue}{chente , it} is a golden \textcolor{blue}{retriver} .    \\

\textbf{EXPECT Target}  &
\textcolor{blue}{i} have a dog \textcolor{red}{and its} name 's \textcolor{blue}{chente , it} is a golden \textcolor{blue}{retriver} .    \\

\hdashline

\textbf{\Dataset{} Source}  &
\textcolor{red}{I} have a dog \textcolor{blue}{its} name 's \textcolor{red}{Chente . It} is a golden \textcolor{red}{retriever} .    \\

\textbf{\Dataset{} Target}  &
\textcolor{red}{I} have a dog \textcolor{red}{and its} name 's \textcolor{red}{Chente . It} is a golden \textcolor{red}{retriever} .    \\

\bottomrule
\end{tabular}}

\caption{
Examples of our \Dataset{}. We mark grammatical errors in \textcolor{blue}{blue} and corrections in \textcolor{red}{red}.
}
\label{tab:expect_examples}
\end{table*}
\begin{table*}[tp!]
\centering
\scalebox{0.75}{
\begin{tabular}{llccc}
\toprule

\textbf{Dataset}  &  &  \textbf{Train}  &  \textbf{Dev}  &  \textbf{Test}  \\
\midrule

\multirow{5}{*}{\textbf{EXPECT}}
&  \textbf{Sentences}         &  15,187  &  2,413  &  2,416  \\
&  \textbf{Sentences with Evidence Words (Perc.)}    &  11,261 (74.15\%)  &  1,426 (59.10\%)  &  1,444 (59.77\%)  \\
&  \textbf{Words / Sentence}      &  28,68  &  29.06  &  29.23  \\ 
&  \textbf{Edits / Sentence}      &  1.03  &  1.08  &  1.07  \\ 
&  \textbf{Evidence Words / Sentence}   &  2.59  &  3.00  &  3.01  \\

\midrule

\multirow{6}{*}{\textbf{\Dataset{}}}
&  \textbf{Sentences}         &  15,187  &  2,413  &  2,416  \\
&  \textbf{Changed Sentences (Perc.)} &  277 (1.82\%)   &  1,311 (54.33\%)  &  1,323 (54.76\%) \\
&  \textbf{Sentences with Evidence Words (Perc.)}    &  11,261 (74.15\%)  &  1,425 (59.06\%)  &  1,443 (59.73\%)  \\
&  \textbf{Words / Sentence}      &  28.52  &  29.53  &  29.72  \\ 
&  \textbf{Edits / Sentence}      &  1.03  &  1.08  &  1.07  \\ 
&  \textbf{Evidence Words / Sentence}  &  2.59  &  3.00  &  3.00  \\

\bottomrule
\end{tabular}}

\caption{
Statistics of the EXPECT and our \Dataset{} datasets. Edits are extracted from the parallel source and target sentences by ERRANT~\citep{bryant-etal-2017-automatic}.
\label{tab:dataset}
}
\end{table*}

\section{\Dataset{} Dataset}
\label{sec:dataset}
Our experiments employ the EXPECT dataset~\citep{fei-etal-2023-enhancing} with human-labeled explanations instead of other LLM-generated explanations~\citep{song-etal-2024-gee}, ensuring the reliability of explanations. The source and target sentences of EXPECT draw from W\&I+LOCNESS~\citep{bryant-etal-2019-bea}, an explanation-free GEC dataset encompassing a broader spectrum of English proficiency levels. EXPECT follows a unique methodology for task simplification. Specifically, for a W\&I+LOCNESS sentence with $n$ grammatical errors, the sentence is replicated $n$ times, each time keeping only one individual error. Considering the complexities of explainable GEC, this approach is sensible and preferred, as it simplifies the task by isolating and extracting evidence for one grammatical error at a time, thereby avoiding the confusion caused by multiple interacting errors within a single sentence.

However, we contend that the original EXPECT dataset includes unidentified grammatical errors, thus biasing the training and the evaluation process. For original sentences with $n$ $(n>1)$ grammatical errors from W\&I+LOCNESS, the authors of EXPECT correct a single error and leave the remaining $n-1$ errors uncorrected, as illustrated in Table~\ref{tab:expect_examples}. These uncorrected errors can confuse models, creating uncertainty about which error to correct and explain, thus complicating model training and evaluation.


To address this issue, we re-correct the previously overlooked grammatical errors while preserving the original single error in EXPECT. This rebuilding process is fully automated. Specifically, we re-correct all uncorrected errors by comparing sentences from EXPECT with those in W\&I+LOCNESS. First, we retrieve the original parallel samples from W\&I+LOCNESS using the open-source toolkit TheFuzz\footnote{\url{https://github.com/seatgeek/thefuzz}}. Then, we identify and correct the underlying grammatical errors using GEC evaluation tools ERRANT~\citep{bryant-etal-2017-automatic} and CLEME~\citep{ye2023cleme}. Notably, during the reconstruction, the grammatical errors, evidence words, and error types in both the EXPECT and \Dataset{} datasets are preserved, except in a few extreme cases (one sample from the development set and one from the test set lack evidence words due to overlapping with the uncorrected errors). In total, 277 (1.82\%), 1,311 (54.33\%), and 1,323 (54.76\%) sentences in our reconstructed training, development, and test sets, respectively, differ from their originals in EXPECT. All these altered samples, initially detected with unidentified grammatical errors, are subsequently re-corrected in \Dataset{}. By ensuring that the evidence words and error types align with the single remaining grammatical error, \Dataset{} facilitates unbiased training and evaluation. Detailed dataset statistics are provided in Table~\ref{tab:dataset}.

\section{Experiments}
\label{sec:experiments}

\subsection{Experimental Settings}
\paragraph{Backbone models.} We employ the pre-trained language models BART-Large~\citep{lewis-etal-2020-bart} and T5-Base~\citep{raffel2020exploring}, in conjunction with the decoder-only Llama3-8B~\citep{dubey2024llama}, as our foundational models. These models have been demonstrated to serve as effective backbones for state-of-the-art GEC models in prior research~\citep{ye-etal-2023-mixedit,zhang-etal-2023-bidirectional,wang-etal-2024-improving-grammatical}. It is worth noting that tokenization may divide a word into multiple BPE units, thereby complicating the extraction of evidence words. Considering that evidence words are typically concise and contiguous, we mandate that the pointer indices encompass all BPE units of the evidence words. In cases where an instance does not contain an evidence word, the target does not predict any pointer index. The specifics of fine-tuning and the hyperparameter configurations are detailed in Appendix~\ref{app:training_details}.

\paragraph{Training settings.}
As outlined in Section~\ref{subsec:model}, we aim to conduct experiments within four distinct training settings using a single unified framework with minimal adjustments. In line with the standard approach, the \textit{Baseline} and \textit{Explanation} settings train models to generate corrections or explanations from input source sentences. In contrast, the \textit{Infusion} models are trained to generate corrections from input source sentences and human-labeled explanations. Significantly, the \textit{Self-Rationalization} setting is further categorized into two sub-settings based on the sequence of generating corrections and explanations: 
1) \textit{pre-explaining} models produce the explanation first, followed by the correction, whereas 2) \textit{post-explaining} models generate the correction first and then the explanation. Generally, we first extract evidence words and subsequently predict error types, as our preliminary experiments indicate that the prediction order of evidence words and error types does not substantially impact performance.

\paragraph{Baselines.}
We regard single-task models as our baseline models. We denote the single-task correction baseline model by \textit{Baseline} and the single-task explanation baseline model by \textit{Explanation}. Additionally, we integrate BERT~\citep{devlin-etal-2019-bert} as an extra explanation baseline. Self-Rationalization models are comparable to the two baseline models since they use only source sentences as input. On the other hand, Infusion models cannot be compared to any baselines since they incorporate extra explanations in their input.

\paragraph{Evaluation.} We evaluate model performance in our experiments from three critical aspects. 1) \textit{Correction}. Following the conventional evaluation setup of W\&I+LOCNESS dataset~\citep{bryant-etal-2019-bea}, we report correction performance using ERRANT~\citep{bryant-etal-2017-automatic}.
2) \textit{Extraction of evidence words}. Following~\citet{fei-etal-2023-enhancing}, we utilize token-level evaluation metrics such as Precision, Recall, F$_1$, and F$_{0.5}$ but exclude the exact match (EM) metric. Interestingly, we observe that \textit{do-nothing} systems achieve higher EM scores than most well-trained systems, yet they still receive zero F$_1$ and F$_{0.5}$ scores, indicating EM may not be a reliable measure in the task. According to~\citet{fei-etal-2023-enhancing}, the F$_{0.5}$ score has the highest correlation with human evaluation (Pearson's coefficient), followed by the F$_1$ score. 3) \textit{Classification of grammatical errors}. We report label accuracy for grammatical error classification performance. Unlike previous studies~\citep{fei-etal-2023-enhancing}, we separate the evaluation of extraction and classification, offering a clearer view of model performance. Specifically, we consider an evidence word a True Positive (TP) if all its BPEs (or their variants) are detected. This approach deviates from previous evaluations~\citep{fei-etal-2023-enhancing}, which classify an evidence word as TP only when both its BPEs and error type are correctly identified. Results are averaged over three runs with different random seeds on the training set, using \Dataset{}-\textit{dev} as the validation set.

\subsection{Comparison of Original and \Dataset{} Datasets}
\label{subsec:exp_datasets}
We first demonstrate the effectiveness of our reconstruction method. To this end, we independently train models respectively on the original EXPECT dataset and its reconstructed version. As shown in Table~\ref{tab:exp_datasets_detailed}, our \Dataset{} dataset significantly enhances performance across both tasks, with the average correction F$_{0.5}$ and explanation F$_{0.5}$ scores increasing by 12.9\% and 4.5\%, respectively. These improvements underscore the high quality of the denoised samples, which result from detecting and rectifying previously overlooked grammatical errors. Consequently, all subsequent experiments will be conducted using the \Dataset{} dataset.

\begin{table*}[t!]
\renewcommand{\arraystretch}{1.2}
\resizebox{1.0\linewidth}{!}{
\begin{tabular}{lcccc}
\toprule

\multicolumn{1}{c}{\multirow{2}{*}{\textbf{System}}}
&  \multicolumn{2}{c}{\textbf{EXPECT-\textit{dev}}}  
&  \multicolumn{2}{c}{\textbf{\Dataset{}-\textit{dev}}}            \\

\cmidrule(lr){2-3} \cmidrule(l){4-5} 
&  \textbf{Cor. (P / R / $\textbf{F}_{0.5}$)$\uparrow$}  
&  \textbf{Exp. (P / R / $\textbf{F}_{1}$ / $\textbf{F}_{0.5}$ / Acc)$\uparrow$}
&  \textbf{Cor. (P / R / $\textbf{F}_{0.5}$)$\uparrow$}  
&  \textbf{Exp. (P / R / $\textbf{F}_{1}$ / $\textbf{F}_{0.5}$ / Acc)$\uparrow$} 
 \\

\hline

\textbf{BART Baseline}  &
30.59 / 33.72 / 31.17  &
-  &
36.14 / 34.87 / \textbf{35.88}  &
-  \\

\hline 

\textbf{Infusion}  \\

\hspace{0.3cm} \textbf{+ Evidence}   &
40.72 / 43.31 / 41.22  &
-  &
45.78 / 44.55 / \textbf{45.53}  &
-  \\

\hspace{0.3cm} \textbf{+ Type}   &
31.15 / 35.14 / 31.87  &
-  &
35.31 / 47.87 / 35.22  &
-  \\

\hspace{0.3cm} \textbf{+ Evidence\&Type}   &
40.79 / 42.50 / 41.11  &
-  &
44.28 / 47.55 / 44.90  &
-  \\

\hline 

\textbf{Self-Rationalization}  \\

\hspace{0.3cm} \textbf{Pre-explaining}  &
32.62 / 31.29 / 32.35  &
33.75 / 44.12 / 38.25 / 35.41 / 28.22  &
38.25 / 34.18 / \textbf{37.36}  &
36.01 / 35.58 / 35.79 / 35.92 / 26.56  \\

\hspace{0.3cm} \textbf{Post-explaining}  &
30.94 / 35.49 / 31.75  &
45.92 / 38.42 / 41.84 / 44.19 / 37.63  &
36.34 / 40.15 / 37.05  &
48.95 / 42.72 / \textbf{45.63} / \textbf{47.56} / \textbf{40.32}  \\

\bottomrule

\end{tabular}}

\caption{Comparison of EXGEC models trained on the original and the \Dataset{} datasets. All experimental results are based on BART-Large.}
\label{tab:exp_datasets_detailed}
\end{table*}

\begin{table*}[tp!]
\renewcommand{\arraystretch}{1.2}

\resizebox{1.0\linewidth}{!}{
\begin{tabular}{llcccc}
\toprule

\multicolumn{1}{c}{\multirow{2}{*}{\textbf{Model}}}
& \multicolumn{1}{c}{\multirow{2}{*}{\textbf{System}}}
&  \multicolumn{2}{c}{\textbf{\Dataset{}-\textit{dev}}}  
&  \multicolumn{2}{c}{\textbf{\Dataset{}-\textit{test}}} \\

\cmidrule(lr){3-4} \cmidrule(l){5-6} 

& 
&  \textbf{Cor. (P / R / $\textbf{F}_{0.5}$)$\uparrow$}  
&  \textbf{Exp. (P / R / $\textbf{F}_{1}$ / $\textbf{F}_{0.5}$ / Acc)$\uparrow$}
&  \textbf{Cor. (P / R / $\textbf{F}_{0.5}$)$\uparrow$}  
&  \textbf{Exp. (P / R / $\textbf{F}_{1}$ / $\textbf{F}_{0.5}$ / Acc)$\uparrow$}  \\

\hline

\textbf{BERT}  &  \textbf{Explanation}  &
-  &
53.60 / 35.46 / \textbf{42.68} / \textbf{48.63} / \textbf{52.09}  &
-  &
51.73 / 36.34 / \textbf{42.69} / \textbf{47.69} / \textbf{50.83}  \\

\hline

\multirow{7}{*}{\textbf{BART}} 
& \textbf{Baseline}  &
36.14 / 34.87 / 35.88  &
-  &
36.33 / 35.49 / 36.16  &
-  \\

& \textbf{Explanation}  &
-  &
44.43 / 32.93 / 37.82 / 41.53 / 33.36  &
-  &
42.34 / 33.13 / 37.18 / 40.11 / 26.95  \\

& \textbf{Infusion + Evidence}   &
45.78 / 44.55 / \textbf{45.53}  &
-  &
46.02 / 44.13 / \textbf{45.63}  &
-  \\

& \textbf{Infusion + Type}   &
35.31 / 47.87 / 35.22  &
-  &
36.00 / 35.37 / 35.87  &
-  \\

& \textbf{Infusion + Evidence\&Type}   &
44.28 / 47.55 / 44.90  &
-  &
44.96 / 47.50 / 45.44  &
-  \\

& \textbf{Self-Rationalization Pre-explaining}  &
38.25 / 34.18 / \textbf{37.36}  &
36.01 / 35.58 / 35.79 / 35.92 / 26.56  &
38.68 / 35.41 / \textbf{37.98}  &
36.77 / 36.85 / 36.81 / 36.79 / 26.24  \\

& \textbf{Self-Rationalization Post-explaining}  &
36.34 / 40.15 / 37.05  &
48.95 / 42.72 / \textbf{45.63} / \textbf{47.56} / \textbf{40.32}  &
36.52 / 40.41 / 37.24  &
49.43 / 44.10 / \textbf{46.61} / \textbf{48.26} / \textbf{39.86}  \\

\hline

\multirow{7}{*}{\textbf{T5}} 
& \textbf{Baseline}  &
44.04 / 22.77 / 37.11  &
-  &
43.31 / 23.71 / 37.16  &
-  \\

& \textbf{Explanation}  &
-  &
58.47 / 24.23 / 34.26 / 45.59 / 28.01   &
-  &
59.00 / 25.13 / 35.25 / 46.47 / 26.06  \\

& \textbf{Infusion + Evidence}   &
52.88 / 31.40 / 46.54  &
-  &
53.50 / 30.76 / 46.61  &
-  \\

& \textbf{Infusion + Type}   &
41.86 / 22.20 / 35.56  &
-  &
44.06 / 23.40 / 37.45  & 
-  \\

& \textbf{Infusion + Evidence\&Type}   &
52.41 / 32.22 / \textbf{46.57}  &
-  &
52.90 / 31.46 / \textbf{46.55}  &
-  \\

& \textbf{Self-Rationalization Pre-explaining}  &
41.60 / 26.13 / 37.19  &
59.87 / 22.78 / 33.01 / 45.16 / 26.61  &
40.96 / 27.20 / 37.19  &
59.33 / 23.27 / 33.43 / 45.29 / 26.03  \\

& \textbf{Self-Rationalization Post-explaining}  &
43.04 / 28.48 / \textbf{39.05}  &
66.82 / 23.95 / \textbf{35.26} / \textbf{49.20} / \textbf{29.34}  &
42.28 / 28.75 / \textbf{38.64}  &
67.99 / 26.10 / \textbf{37.72} / \textbf{51.47} / \textbf{29.80}  \\

\hline

\multirow{7}{*}{\textbf{Llama3}} 
& \textbf{Baseline}  &
33.80 / 37.53 / 34.48  &
-  &
33.64 / 38.12 / 34.45  &
-  \\

& \textbf{Explanation}  &
-  &
47.70 / 6.53 / 11.49 / 21.10 / 23.17  &
-  &
54.68 / 7.39 / 13.02 / 23.98 / 27.32  \\

& \textbf{Infusion + Evidence}   &
40.15 / 44.47 / 40.95  &
-  &
41.94 / 46.26 / 42.74  &
-  \\

& \textbf{Infusion + Type}   &
37.35 / 42.77 / 38.32 &
-  &
40.20 / 44.60 / 41.01  &
-  \\

& \textbf{Infusion + Evidence\&Type}   &
48.01 / 52.64 / \textbf{48.87}  &
-  &
45.26 / 49.75 / \textbf{46.09}  &
-  \\

& \textbf{Self-Rationalization Pre-explaining}  &
32.55 / 36.72 / 33.31  &
53.84 / 8.68 / 14.94 / 26.39 / 27.35  &
34.98 / 38.36 / \textbf{35.61} &
55.51 / 10.31 / 17.39 / 29.58 / 28.89  \\

& \textbf{Self-Rationalization Post-explaining}  &
32.37 / 41.08 / \textbf{33.80}  &
68.97 / 29.29 / \textbf{41.12} / \textbf{54.27} / \textbf{36.84}  &
33.58 / 41.84 / 34.96  &
64.36 / 27.34 / \textbf{38.38} / \textbf{50.64} / \textbf{35.22}  \\

\bottomrule
\end{tabular}}

\caption{Main results on \Dataset{}-\textit{dev} and \Dataset{}-\textit{test}.}
\label{tab:exp_main}

\end{table*}

\subsection{Main Results}
Here, we inspect and evaluate the relationship between the correction and the explanation tasks through various experimental training setups. The main results presented in Table~\ref{tab:exp_main} reveal the following insights.

\paragraph{Evidence words offer more valuable information than grammatical error types for corrections.}
First, we investigate the Infusion setting, where we append different supplementary explanation data to the input source. Infusion models serve as an oracle since human-annotated explanations are typically lacking in real-world applications, allowing us to understand the impact of explanations on the correction task. Recent research has demonstrated that adding human-annotated explanations as additional input can improve task performance~\citep{hase-etal-2020-leakage,yao-etal-2023-human}, and we have observed similar results in the \textit{Infusion} setting. Specifically, we find that the additional information provided by grammatical error types enhances the correction performance of Llama3, but does not positively affect the correction performance of BART and T5. We suspect the significant enhancement of Llama3's correction performance is due to that Llama3 can understand the semantic meaning of error types. In contrast, offering evidence words can consistently boost the correction F$_{0.5}$ scores by 5$\sim$20 points for three different language models, even though only about 60\% of the samples in the dev and test sets are annotated with evidence words, showing that accurate evidence words are very beneficial for the correction task.

\paragraph{Jointly learning the two tasks is advantageous.}
In practical applications, explanations are generally unavailable during inference. Thus, Self-Rationalization models investigate whether incorporating explanations as additional output can enhance performance. The Self-Rationalization models' correction F$_{0.5}$ scores improve by an average of 1.3 points for BART and 0.88 points for T5 compared to the \textit{Baseline} setting. Similarly, explanation F$_{0.5}$ scores increase by an average of 1.3 points for BART, 1.8 points for T5, and 17.7 points for Llama3, compared to the \textit{Explanation} setting. The error classification accuracy scores see an average improvement of 9.8 points for BART, 0.9 points for T5, and 6.8 points for Llama3.

\paragraph{Post-explaining models perform better in the explanation performance.} Experiments reveal variations in performance between models that explain beforehand and those that explain afterward across three language models. We observe that post-explaining models predict evidence words and error types more accurately. This could be attributed to the challenge of predicting evidence words and error types without a specific correction. Furthermore, pre-explaining models exhibit marginally better correction performance for BART, contrasting with the trends observed in other models.


Furthermore, a BERT model based on sequence labeling is trained under consistent training and evaluation conditions from~\citet{fei-etal-2023-enhancing}. The generative language models exhibit significantly lower performance in grammatical error type classification compared to BERT-based models, which we hypothesize is due to intrinsic biases introduced by the differences in auto-regressive text generation and BERT's masked language model (MLM) pre-training objectives. This hypothesis is supported by the experiments in Section~\ref{subsec:exp_tag}, indicating that sequence labeling is not essential for grammatical error type classification. We report detailed results of EXGEC models in Appendix~\ref{app:detailed_results} and provide an investigation of varying loss weights $\lambda$ in Appendix~\ref{app:loss_weight}.

\section{Analyses}
\label{sec:analyses}
In this section, we adopt BART-Large for further analyses, aiming to provide insights into our framework design and the effects of explanations.

\subsection{Does Sequence Labeling Help?}
\label{subsec:exp_tag}
Motivated by recent studies in multi-task GEC frameworks~\citep{zhao-etal-2019-improving,yuan-etal-2021-multi}, which combine a sequence labeling task with a sentence-level correction task, we also develop a multi-task baseline for explainable GEC, keeping the experimental setup the same as our other experiments. Specifically, we append a randomly initialized tagging head after the encoder to perform the explanation task as a sequence labeling task, like BERT-based models. To predict each token's tag, we pass the encoder's hidden representation $\mathbf{H}^e$ through a softmax after an affine transformation, which is computed as follows:
\begin{equation}\begin{aligned}
P(T \mid X) = \operatorname{softmax}(W^{\top} \mathbf{H}^e),
\end{aligned}\end{equation}
\noindent where $T$ denotes the tagging sequence in the BIO scheme. To replace the pointer mechanism, we employ a token-level sequence labeling task, focusing solely on the correction task in the decoder. We also implement loss weighting to balance the correction generation and sequence labeling losses, defined as follows:
\begin{equation}\begin{aligned}
\mathcal{L} = \mathcal{L}_{cor} + \gamma \cdot \mathcal{L}_{tag},
\end{aligned}\end{equation}
\noindent where $\gamma$ represents the trade-off factor, and we minimize the cross-entropy between predicted tokens/labels and ground truth tokens/labels.

The outcomes of varying $\gamma$, chosen from the alternative set $\{0.5,0.8,1.0,1.5,2.0\}$, are presented in Table~\ref{tab:exp_tag}. Compared to Self-Rationalization models, sequence labeling-based multi-task models demonstrate inferior correction performance; however, they provide an intermediary level of explanation performance between pre-explaining and post-explaining models. So it can be inferred that our proposed EXGEC exhibits greater efficacy than sequence labeling-based baselines.


\begin{table}[tp!]
\renewcommand{\arraystretch}{1.3}

\centering
\resizebox{\linewidth}{!}{
\begin{tabular}{lcc}
\toprule

$\gamma$  &
\textbf{Cor. (P / R / $\textbf{F}_{0.5}$)$\uparrow$}  &
\textbf{Exp. (P / R / $\textbf{F}_{1}$ / $\textbf{F}_{0.5}$ / Acc)$\uparrow$}  \\

\midrule

0.5  &  36.16 / 35.68 / 36.06  &
57.00 / 06.87 / 12.26 / 23.18 / 19.15  \\

0.8  &  35.47 / 36.92 / 35.74  &
51.77 / 21.63 / 30.51 / 40.49 / 23.46  \\

1.0  &  35.10 / 36.96 / 35.46  &
48.82 / 26.55 / \textbf{34.40} / \textbf{41.81} / 25.94  \\

1.5  &  36.12 / 36.34 / \textbf{36.16}  &
50.95 / 22.01 / 30.74 / 40.34 / 24.66  \\

2.0  &  35.93 / 35.38 / 35.82  &
52.48 / 22.29 / 31.29 / 41.29 / \textbf{28.06}  \\

\bottomrule

\end{tabular}}

\caption{Results of sequence labeling-based multi-task BART baselines for varying loss weights $\gamma$ on \textbf{\Dataset{}-\textit{dev}}.}
\label{tab:exp_tag}

\end{table}
\begin{table*}[tp!]
\renewcommand{\arraystretch}{1.2}
\centering
\resizebox{1.0\linewidth}{!}{
\begin{tabular}{lcccc}
\toprule

\multicolumn{1}{c}{\multirow{2}{*}{\textbf{System}}}
&  \multicolumn{2}{c}{\textbf{\Dataset{}-\textit{dev}}}  
&  \multicolumn{2}{c}{\textbf{\Dataset{}-\textit{test}}} \\

\cmidrule(lr){2-3} \cmidrule(l){4-5} 
&  \textbf{Cor. (P / R / $\textbf{F}_{0.5}$)$\uparrow$}  
&  \textbf{Exp. (P / R / $\textbf{F}_{1}$ / $\textbf{F}_{0.5}$ / Acc)$\uparrow$}
&  \textbf{Cor. (P / R / $\textbf{F}_{0.5}$)}  
&  \textbf{Exp. (P / R / $\textbf{F}_{1}$ / $\textbf{F}_{0.5}$ / Acc)$\uparrow$}

\\

\hline

\textbf{Baseline}  &
36.14 / 34.87 / 35.88  &
-  &
36.33 / 35.49 / 36.16  &
-  \\

\hline

\textbf{Infusion}  \\

\hspace{0.3cm} \textbf{+ G.T. Evidence}   &
45.78 / 44.55 / \textbf{45.53}  &
-  &
46.02 / 44.13 / \textbf{45.63}  &
-  \\

\hspace{0.3cm} \textbf{+ Ran. Evidence}   &
35.88 / 33.26 / 35.33  &
-  &
36.44 / 33.20 / 35.74  &
-  \\

\hspace{0.3cm} \textbf{+ Adj. Evidence}   &
38.46 / 42.81 / 39.26  &
-  &
39.66 / 43.01 / 40.28  &
-  \\

\hline 

\textbf{Pre-explaining}  \\

\hspace{0.3cm} \textbf{+ G.T. Evidence}  &
38.25 / 34.18 / \textbf{37.36}  &
36.01 / 35.58 / \textbf{35.79} / \textbf{35.92} / \textbf{26.56}  &
38.68 / 35.41 / \textbf{37.98}  &
36.77 / 36.85 / \textbf{36.81} / \textbf{36.79} / \textbf{26.24}  \\

\hspace{0.3cm} \textbf{+ Ran. Evidence}  &
36.17 / 33.72 / 35.65  &
13.60 / 00.40 / 00.77 / 01.79 / 15.83  &
37.63 / 34.83 / 37.04  &
14.38 / 00.53 / 01.02 / 02.31 / 15.02  \\

\hspace{0.3cm} \textbf{+ Adj. Evidence}  &
36.53 / 38.73 / 36.95  &
26.97 / 03.37 / 06.00 / 11.23 / 17.03  &
37.09 / 39.52 / 37.55  &
29.00 / 04.02 / 07.06 / 12.93 / 16.02  \\

\hline 

\textbf{Post-explaining}  \\

\hspace{0.3cm} \textbf{+ G.T. Evidence}  &
36.34 / 40.15 / \textbf{37.05}  &
48.95 / 42.72 / \textbf{45.63} / \textbf{47.56} / \textbf{40.32}  &
36.52 / 40.41 / \textbf{37.24}  &
49.43 / 44.10 / \textbf{46.61} / \textbf{48.26} / \textbf{39.86}  \\

\hspace{0.3cm} \textbf{+ Ran. Evidence}  &
36.36 / 34.37 / 35.95  &
14.39 / 00.45 / 00.86 / 02.00 / 16.04  &
36.86 / 34.87 / 36.44  &
07.45 / 00.16 / 00.32 / 00.74 / 15.02  \\

\hspace{0.3cm} \textbf{+ Adj. Evidence}  &
36.36 / 34.14 / 35.89  &
23.68 / 02.53 / 04.57 / 08.86 / 15.79  &
37.34 / 35.18 / 36.88  &
26.74 / 03.28 / 05.84 / 11.00 / 15.48  \\

\bottomrule

\end{tabular}}
\caption{Results of BART-Large models trained on \textit{Ground Truth} (G.T.), \textit{Random} (Ran.) or \textit{Adjacent} (Adj.) evidence words.}
\label{tab:exp_synthesize}
\end{table*}

\subsection{Position Leakage}
One might suspect that the improved performance of Infusion models is driven, at least in part, by a positional leakage effect—where evidence words often appear within the first- or second-order nodes of correction words in the dependency parsing tree~\citep{fei-etal-2023-enhancing}. To address this concern, we synthesize datasets with artifact explanations using two methods: (1) \textit{random explanations}, which are randomly sampled from the entire set of source tokens, and (2) \textit{adjacent explanations}, which are randomly chosen from candidate source tokens located within a distance of 1 to 5 tokens from the correction position (noting that over 90\% of evidence words in EXPECT fall within this range).

Given that many samples lack annotated evidence words, we generate a number of synthesized evidence words matching the count of the ground-truth annotations to ensure experimental fairness. While our models are trained using these synthesized evidence words, evaluation is conducted with the ground-truth evidence words. This setup allows us to investigate whether the models can effectively learn to extract evidence words through an unsupervised approach. The results of these experiments are presented in Table~\ref{tab:exp_synthesize}.

Under the Infusion setting, it is unsurprising that incorporating random evidence words does not enhance correction performance as we anticipated. However, we observe that using adjacent synthesized evidence words produces a noticeable impact—yielding a moderate improvement over random evidence words, albeit still falling short of the gains achieved with ground-truth evidence words. This indicates that a positional leakage effect does exist, although it alone cannot fully realize the benefits of ground-truth evidence.

In the pre-explaining and post-explaining settings, generating adjacent evidence words yields a modest improvement in correction performance. However, this gain remains inferior to the enhancement achieved when incorporating ground-truth evidence words, underscoring the critical importance of jointly training the correction and explanation tasks. In contrast, employing random evidence words offers no discernible benefit for the correction task, and further evaluations reveal that their contributions are largely disregarded in the model’s explanations. Notably, the absence of ground-truth evidence words results in a marked decline in explanation performance—reflecting the inherent challenge of aligning model-generated explanations with human preferences in an unsupervised context.

These findings emphasize that high-quality and contextually relevant evidence is essential for improving correction accuracy and producing explanations that resonate with human evaluators.

\section{Limitations}
\label{sec:limitations}

\paragraph{The limited nature of EXPECT.}
The explanations provided in the EXPECT and \Dataset{} datasets are confined to simple evidence words and types of grammatical errors. These may not be intuitive or comprehensive for L2 speakers, who are usually the primary users of educational GEC systems. However, our experiments demonstrate that the explanations can significantly aid the correction task by effectively utilizing the \Dataset{} dataset. In our future work, we aim to investigate more general free-text explanations in real-world GEC applications, presenting an encouraging direction for enhancing user-focused GEC systems.

EXPECT and \Dataset{} suppose only one grammatical error exists in a sentence, which is a less realistic setting. Due to the undeveloped research on the explainability of GEC, there are few high-quality datasets to study. Focusing on the \Dataset{} dataset, our experiments follow the setting even though the proposed EXGEC can flexibly adapt to other real-life settings.

\paragraph{Inherent nature of generative language models.}
It has been observed that the backbone models we employ, specifically BART, T5, and Llama3, exhibit deficiencies in classifying grammatical errors when compared to BERT-based models. This shortcoming can be ascribed to their fundamental characteristics as generative language models. Such limitations may adversely affect correction performance, especially in post-explanatory models that rectify sentences by relying on previously forecasted explanations. Our future research intends to explore more effective methodologies for managing and integrating both these tasks.



\section{Conclusion}
\label{sec:conclusion}

This paper aims to improve the explainability of GEC systems. To this end, we introduce \Method{}, a unified generative framework that simultaneously performs correction and explanation tasks. The framework empirically explores and establishes the interaction between the two tasks. We then identify the potential drawback of EXPECT and propose \Dataset{}. Results on three language models reveal performance enhancement when multi-task learning correction and explanation tasks. In the future, we plan to extend our framework to free-text explanations.

\bibliography{tacl2021,anthology}
\bibliographystyle{acl_natbib}

\appendix
\begin{table*}[!tb]
\renewcommand{\arraystretch}{1.2}
\centering

\resizebox{1.0\linewidth}{!}{
\begin{tabular}{lcccc}
\toprule

\textbf{System}
&  \textbf{Cor. (TP / FP / FN)}
&  \textbf{Cor. (P / R / $\textbf{F}_{0.5}$)$\uparrow$}
&  \textbf{Exp. (TP / FP / FN)}
&  \textbf{Exp. (P / R / $\textbf{F}_{1}$ / $\textbf{F}_{0.5}$ / Acc)$\uparrow$} \\

\hline

\textbf{BART Baseline} &
910 / 1604 / 1695  &
36.14 / 34.87 / 35.88  &
-  &
-  \\

\hline

\textbf{Infusion}  \\

\hspace{0.3cm} \textbf{+ Evidence}   &
1149 / 1345 / 1459  &
45.78 / 44.55 / \textbf{45.53}  &
-  &
-  \\

\hspace{0.3cm} \textbf{+ Type}   &
879 / 1608 / 1716  &
35.31 / 47.87 / 35.22  &
-  &
-  \\

\hspace{0.3cm} \textbf{+ Evidence\&Type}   &
1244 / 1600 / 1351  &
44.28 / 47.55 / 44.90  &
-  &
-  \\

\hline 

\textbf{Self-Rationalization}  \\

\hspace{0.3cm} \textbf{Pre-explaining}  &
885 / 1437 / 1721  &
38.25 / 34.18 / \textbf{37.36}  &
1525 / 2701 / 2737  &
36.01 / 35.58 / 35.79 / 35.92 / 26.56  \\

\hspace{0.3cm} \textbf{Post-explaining}  &
1038 / 1821 / 1548  &
36.34 / 40.15 / 37.05  &
1829 / 1911 / 2456  &
48.95 / 42.72 / \textbf{45.63} / \textbf{47.56} / \textbf{40.32}  \\

\bottomrule

\end{tabular}}

\caption{Detailed results of BART-Large on \Dataset{}-\textit{dev}, including the number of True Positive (TP), False Positive (FP), and False Negative (FP) for both correction and explanation tasks. TP / FP / FN counts are taken from one checkpoint, while P / R / F / Acc scores are averaged over three runs.}
\label{tab:detailed_results}
\end{table*}

\begin{table*}[!tbh]
\centering
\renewcommand{\arraystretch}{1.2}
\label{tab:hp}
\scalebox{0.80}{
\begin{tabular}{clcccc}
\toprule

& \textbf{Configuration}  & \textbf{BART}  & \textbf{T5}  & \textbf{LLaMA3} \\ 
\midrule

\multirow{9}{*}{\textbf{Training}}
& Backbone  & BART-Large & T5-Base & LLaMA3-8B  \\
& Epochs    & 60 & 80 & 3 (LoRA) \\
& Batch size per GPU & 4096 tokens & 4096 tokens & 8192 tokens \\
& Gradient Accumulation & 4 & 4 & 8 \\

\cdashline{2-5}

& Loss weight $\lambda$& \multicolumn{3}{c}{ 1.0 } \\
& Learning rate &  \multicolumn{3}{c}{$3 \times 10^{-5}$ } \\
& Devices   & \multicolumn{3}{c}{1 Tesla A100 GPU (80GB)} \\
& \multirow{2}{*}{Optimizer} & \multicolumn{3}{c}{Adam \citep{kingma2014adam}}  \\
& & \multicolumn{3}{c}{($\beta_1=0.9,\beta_2=0.999,\epsilon=1\times 10^{-8}$) } \\

\midrule

\multirow{2}{*}{\textbf{Inference}}
& Beam size          & 5 & 5 & 5 \\
& Max length         & 256 & 512 & 512 \\

\bottomrule
\end{tabular}}
\caption{Hyper-parameters used in our experiments.}
\label{tab:hyper-parameter}
\end{table*}

\begin{table}[tp!]
\renewcommand{\arraystretch}{1.3}
\renewcommand{\tabcolsep}{10pt}

\resizebox{\linewidth}{!}{
\begin{tabular}{lcc}
\toprule

$\lambda$  &
\textbf{Cor. (P / R / $\textbf{F}_{0.5}$)$\uparrow$}  &
\textbf{Exp. (P / R / $\textbf{F}_{1}$ / $\textbf{F}_{0.5}$ / Acc)$\uparrow$}  \\

\midrule

0.5  &  35.40 / 38.03 / 35.90  &
39.77 / 38.88 / 39.32 / 39.59 / 32.02  \\

1.0  &  36.34 / 40.15 / \textbf{37.05}  &
48.95 / 42.72 / \textbf{45.63} / \textbf{47.56} / \textbf{40.32}  \\

1.5  &  36.03 / 38.42 / 36.49  &
43.90 / 42.82 / 43.35 / 43.68 / 36.88  \\

2.0  &  35.41 / 38.61 / 36.00  &
47.98 / 42.86 / 45.28 / 46.86 / 40.07  \\

\bottomrule

\end{tabular}}
\caption{Results of \textit{post-explaining} models with varying loss weights $\lambda$ on \textbf{\Dataset{}-\textit{dev}}.}
\label{tab:exp_loss}
\end{table}

\section{Fine-tuning Details}
\label{app:training_details}
We update all parameters of BART-Large and T5-Base during fine-tuning. For Llama3-8B, we leverage LoRA~\citep{hu2022lora} to implement parameter-efficient fine-tuning. Detailed hyper-parameter configurations are provided in Table~\ref{tab:hyper-parameter}. 

\section{Detailed Results}
\label{app:detailed_results}
Table~\ref{tab:detailed_results} lists detailed results on the \Dataset{}-\textit{dev} set, providing further insight into model behaviors in different training settings. Infusion (with Evidence and with Evidence \& Type) models gain higher correction TP but lower FP and FN, demonstrating that evidence words significantly benefit the correction task. Additionally, pre-explaining models tend to extract more evidence words ($\approx$ 4200\footnote{This is equal to TP plus FP.}) but predict fewer correction edits ($\approx$ 2300).
On the other hand, post-explaining models are declined to extract fewer evidence words ($\approx$ 3700), but predict more correction edits ($\approx$ 2900). We speculate that models are more likely to make predictions when prior information is unavailable. However, models become more cautious with the available prior information. Therefore, pre-explaining models achieve higher correction precision, whereas post-explaining models exhibit higher correction recall.

\section{Impact of Loss Weighting}
\label{app:loss_weight}
This section examines the balance between learning for correction and explanation tasks by adjusting the loss weight $\lambda$ in Equation~\eqref{eq:loss}. An elevated value of $\lambda$ indicates an increased focus on learning the explanation task. We experiment with various models employing different loss weights $\lambda$ selected from $\{0.5, 1.0, 1.5, 2.0\}$, and present our results in \Dataset{}-$\textit{dev}$ in Table~\ref{tab:exp_loss}. The findings indicate that focusing predominantly on a single task may cause a deterioration in the performance of both tasks. We hypothesize that with a reduced $\lambda$, the weak supervisory signals in the explanations hinder the model's capacity to generate accurate explanations, thereby compromising the correction learning process. Conversely, a larger $\lambda$ might neglect correction learning, leading to subpar explanation performance since the quality of explanations depends on the predicted corrections in post-explaining models. An optimal $\lambda$ must be chosen to attain a balance between the learning of both tasks and achieve mutually beneficial results.







  

\end{document}